\pdfoutput=1

\documentclass[11pt]{article}
\usepackage[table]{xcolor}
\usepackage{ACL2023}

\usepackage{booktabs}
\usepackage{makecell}

\usepackage{times}
\usepackage{latexsym}
\usepackage{amsmath, amssymb}
\usepackage{amsfonts}
\usepackage{multirow}
\usepackage{rotating}
\usepackage{booktabs}
\usepackage{amsmath}

\usepackage{graphicx}
\usepackage{subcaption}
\usepackage{mwe}
\usepackage{hyperref} 
\usepackage{algorithm}
\usepackage[noend]{algpseudocode}

\usepackage[T1]{fontenc}

\usepackage[utf8]{inputenc}

\usepackage{microtype}

\usepackage{inconsolata}

\title{MARS: Meaning-Aware Response Scoring \\
for Uncertainty Estimation in Generative LLMs }

\author{Yavuz Faruk Bakman \\
  USC \\
  \texttt{ybakman@usc.edu} \\\And
   Duygu Nur Yaldiz \\
  USC \\
  \texttt{yaldiz@usc.edu} \\\And
  Baturalp Buyukates \\
  USC \\
  \texttt{buyukate@usc.edu} \\ \AND 
  Chenyang Tao\footnotemark[1] \\
  Amazon AI \\
  \texttt{chenyt@amazon.com} \\\And
    Dimitrios Dimitriadis\footnotemark[1] \\
  Amazon AI \\
  \texttt{dbdim@amazon.com} \\\And
   Salman Avestimehr \\
  USC \\
  \texttt{avestime@usc.edu}}

\begin{document}

\renewcommand*{\thefootnote}{\fnsymbol{footnote}}

\maketitle
\begin{abstract}
Generative Large Language Models (LLMs) are widely utilized for their excellence in various tasks. However, their tendency to produce inaccurate or misleading outputs poses a potential risk, particularly in high-stakes environments. Therefore, estimating the correctness of generative LLM outputs is an important task for enhanced reliability. 
Uncertainty Estimation (UE) in generative LLMs is an evolving domain, where SOTA probability-based methods commonly employ length-normalized scoring. In this work, we propose Meaning-Aware Response Scoring (MARS) as an alternative to length-normalized scoring for UE methods. MARS is a novel scoring function that considers the semantic contribution of each token in the generated sequence in the context of the question. 
We demonstrate that integrating MARS into UE methods results in a universal and significant improvement in UE performance. We conduct experiments using three distinct closed-book question-answering datasets across five popular pre-trained LLMs. Lastly, we validate the efficacy of MARS on a Medical QA dataset. Code can be found \href{https://github.com/Ybakman/LLM_Uncertainty} {here}. 

\end{abstract}
\footnotetext[1]{This work does not relate to their position at Amazon.}

\section{Introduction}


Generative Large Language Models (LLMs) have risen in popularity due to their remarkable ability to understand, generate, and process human language at an unprecedented scale and accuracy \cite{ye2023comprehensive,openai2023gpt4, touvron2023llama2}. These models have become the state-of-the-art in various fields, including  machine translation, content generation, and even scientific research \cite{huang2023benchmarking, openai2023gpt4} due to their capability to handle diverse tasks such as text summarization, sentiment analysis, and question-answering in a few-shot or zero-shot manner.


\begin{figure*}[!htbp]
\begin{center}
\includegraphics[width=0.9\textwidth]{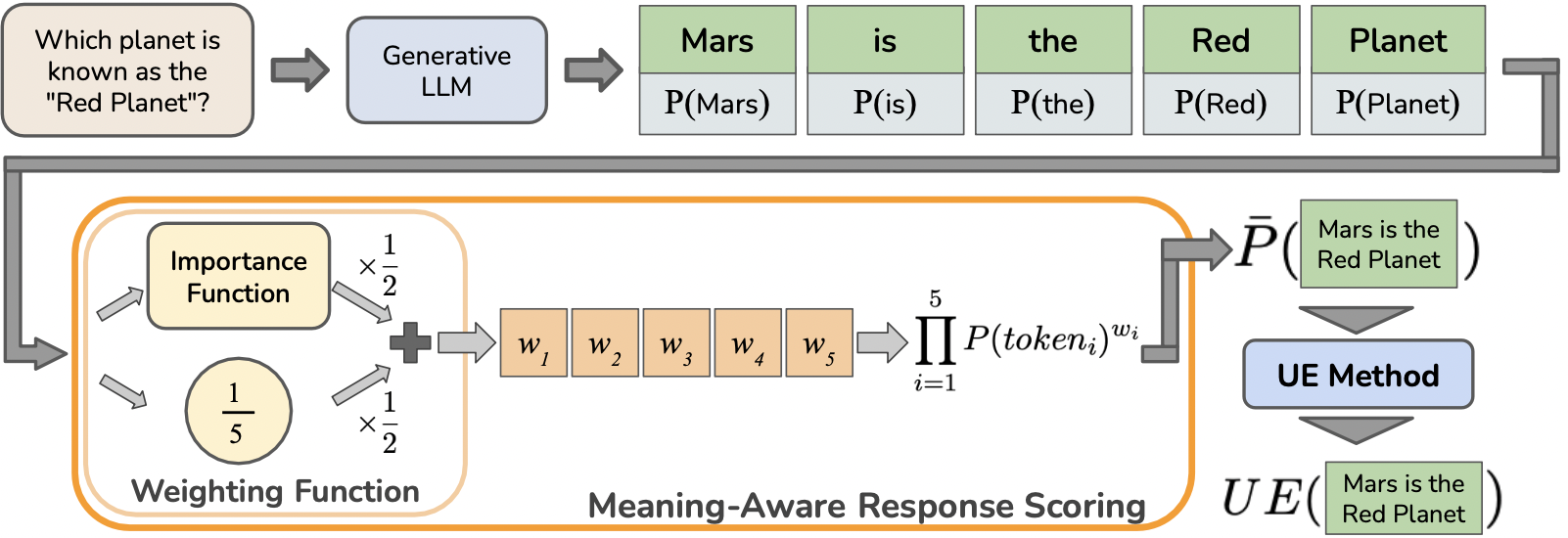}
\caption{Overview of Meaning-Aware Response Scoring (MARS). Each token in the response of a generative LLM is assigned a weight based on its importance in the meaning. The product of the weighted probabilities of these tokens yields the response score. MARS is then used for Uncertainty Estimation (UE) methods in generative LLMs.}
\label{fig:mars}
\end{center}
\vskip -0.2in
\end{figure*}


Despite their growing popularity and success, generative LLMs are not infallible and can sometimes produce erroneous or misleading outputs, especially when dealing with complex reasoning problems or closed-book questions. This limitation becomes particularly critical in question-answering systems used in high-stakes environments. Quantifying the uncertainty of generative LLM responses in such scenarios is not just beneficial but essential for ensuring trustworthy operation. For example, in a medical advice application, accurately assessing the uncertainty of the responses provided by LLMs can prevent the provision of incorrect medical advice. This is crucial because erroneous advice may lead to devastating medical missteps or misunderstandings. Thus, understanding and quantifying uncertainty helps in reliable risk assessment and in maintaining the overall quality of the answers provided, ensuring that users can assess how much reliance they should place on LLM responses.

Uncertainty Estimation (UE) is a well-studied problem in classification scenarios, especially in the computer vision domain \cite{lakshminarayanan2017simple, gal2016dropout,shen2021real}. The proposed UE methods in classification tasks, which rely on the class probabilities, are not directly applicable to generative LLMs due to the auto-regressive generative structure of LLMs \cite{malinin2021uncertainty}, which implies that LLMs generate text sequentially by predicting each subsequent word based on the combined context of all preceding words. This process differs significantly from classification tasks, where the output is typically a single label or a set of labels assigned to an entire input, without the sequential and context-accumulating nature of generative LLMs. 

Recent work \cite{malinin2021uncertainty}, formalizes how to adapt popular UE methods developed for classification tasks
to the context of generative LLMs. They propose using length-normalized scoring to estimate the likelihood of a sequence generated by the model, and the subsequent works \cite{kuhn2023semantic, lin2023generating, chen2023quantifying} utilize that idea of length-normalized scoring.


A downside of these existing UE techniques in the generative LLM literature is treating length-normalized scoring like the class probabilities in classification tasks.
However, better ways may exist for estimating uncertainty than directly using the length-normalized score of a sequence, as it treats all tokens equally.
In reality, each word's contribution to the sentence's meaning in the question context might vary. For example, given the question ``Which planet is known as the Red Planet?'' and with the generated response ``Mars is the Red Planet'', the tokens of ``Mars'' are the most critical ones in the response because those tokens are the ones actually answering the question. Thus, assigning more weight to semantically significant tokens in the response score calculation can improve UE methods, resulting in more accurate predictions.



Based on this word importance intuition, we propose a novel scoring function for generative LLMs called \emph{Meaning-Aware Response Scoring (MARS),} as outlined in Figure~\ref{fig:mars}.
To compute the LLM response score as an input to UE methods, we first assign an importance coefficient to each token in the generation. This importance essentially reflects the impact of masking a token in a sequence on the meaning of the generated response, where tokens with a greater influence on the meaning receive higher importance.
By leveraging these meaning-aware coefficients ($w_i$ in Figure~\ref{fig:mars}), MARS returns the multiplication of the weighted probabilities of the tokens in the generated sequence. 

We list our main contributions as follows:
\vskip -0.1in
\begin{itemize}
\itemsep -0.05in
    \item We propose a novel scoring function for UE in generative LLMs named Meaning-Aware Response Scoring (MARS).
     
    \item We introduce a BERT-like model, efficiently assigning meaning-aware importance weights to the tokens in a single model pass within MARS calculation.

    \item We explain previous works' \cite{malinin2021uncertainty, kuhn2023semantic} design choices from the classification perspective to create a grounded framework for MARS and other scoring functions.
    
    \item We evaluate probability-based UE metrics with MARS on question-answer datasets and show that MARS universally improves the UE performance for an extensive list of LLMs.
\end{itemize}


\section{Background}
In this section, we will go over probability-based UE methods that our work built on. For a detailed discussion on related works, refer to Appendix \ref{related-work}.

In the literature, UE is used as a proxy for the correctness of the model output \cite{malinin2021uncertainty, gal2016dropout, lakshminarayanan2017simple, band2022benchmarking}. For generative LLMs in the question-answer context, we consider the most probable sequence as the model output and utilize UE to predict the correctness of the response following \citet{kuhn2023semantic}. The goal of UE is to assign higher scores to incorrect responses, indicating greater uncertainty, and lower scores to correct responses, signifying less uncertainty.

\subsection{Bayesian View to Estimate Uncertainty}
Bayesian UE is used in machine learning to quantify uncertainty in predictions. 
It treats model parameters as random variables, assigning a prior probability distribution to them. Through Bayesian inference, this distribution is updated with training data, yielding a posterior distribution. Prediction uncertainty stems from this posterior distribution.
Let $\{\theta_i\}^M_{i=1}$ be an ensemble of models sampled from approximate posterior $ q(\theta) \approx p(\theta | D)$ where $D$ is the training data. 

The predictive posterior of input ${x} \in \mathcal{X} $ for target ${y} \in \mathcal{Y} $ is derived by expectation over the ensemble:
\begin{equation}
\begin{split}
P({y}|{x}, D) &= \mathbb{E}_{q(\theta)} [P({y}|{x}, \theta)] \\
          &\approx \frac{1}{M} \sum_{m=1}^{M} P({y}|{x}, \theta_m),
\end{split}\label{posterior}
\end{equation}
where we have $\theta_m \sim q(\theta) \approx p(\theta | D)$.
Using the posterior probability definition, we can define the entropy of predictive posterior as:
\begin{equation}
\mathcal{H}({x},D)= - \sum_{{y} \in \mathcal{Y}} P({y}|{x}, D) \log P({y}|{x}, D).
\end{equation}

In classification tasks, commonly used tools for estimating uncertainty are the entropy of the predictive posterior and the negative predictive posterior probability of the most probable answer \cite{gal2016dropout, lakshminarayanan2017simple, malinin2021uncertainty,xiao2022uncertainty,chen2023quantifying}.
However, the formulation in (\ref{posterior}) is not applicable to generative LLMs because of their auto-regressive generative structure.

\subsection{Uncertainty Estimation (UE) of Auto-Regressive Generative Models}

\begin{figure*}[!htbp]
\begin{center}
\includegraphics[width=0.9\textwidth]{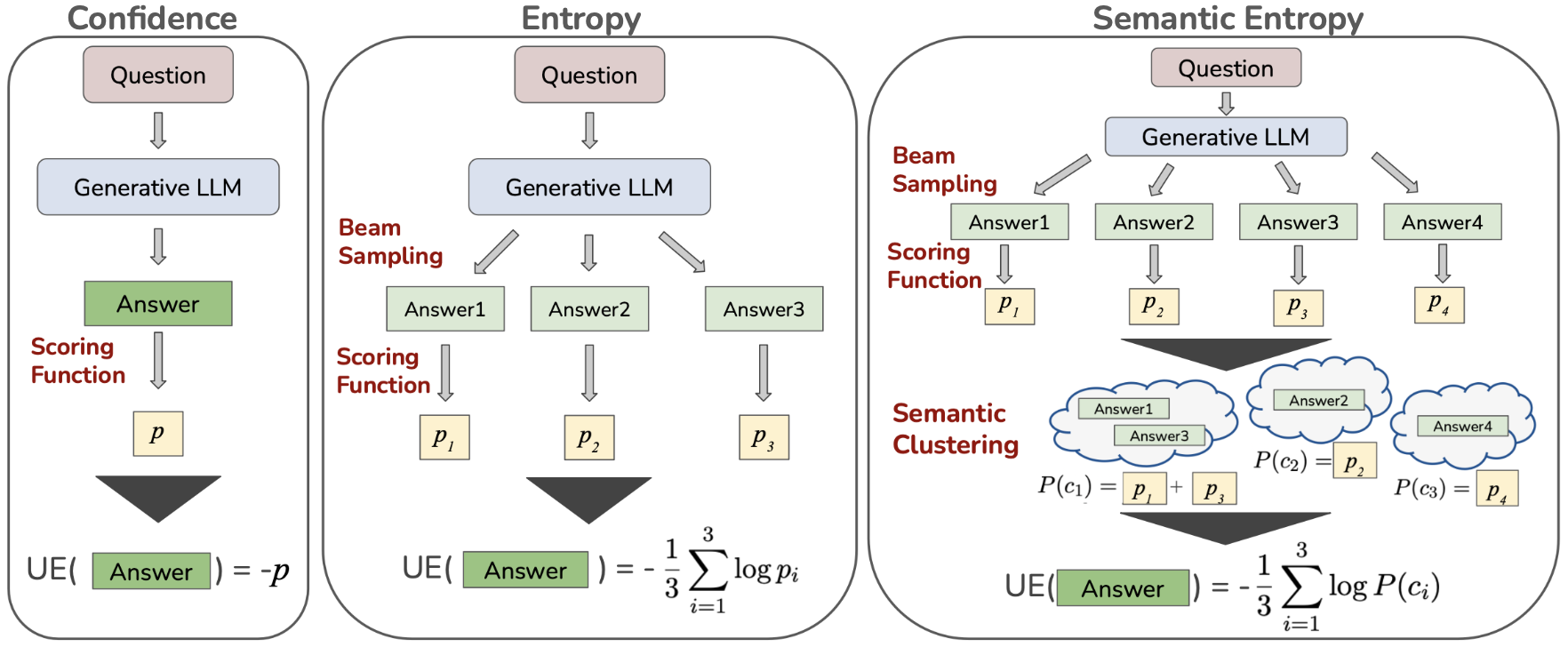}
\caption{The most common probability-based UE methods for generative LLMs. The aim is to calculate the uncertainty of the most probable answer (shown in darker green) to the given question. Length-normalized scoring (\ref{length-normalized-prob}) is used in all these methods to obtain output scores. We propose MARS to replace it in these schemes.}
\label{fig:background}
\end{center}
\vskip -0.2in
\end{figure*}

\citet{malinin2021uncertainty} formalizes posterior probability definition for auto-regressive generative models where the output $\mathbf{s}$ is not a single entity but a sequence of tokens $ \mathbf{s}  = \{s_1, s_2, ..., s_L \}$. They simply replace $P({y}|{x}, \theta)$ in (\ref{posterior}) with sequence probability $P(\mathbf{s}|\mathbf{x}, \theta)$. The probability of a sequence $\mathbf{s}$ for a given model parametrized with $\theta$ is defined as the multiplication of probabilities of its tokens: 
\begin{equation}
    P(\mathbf{s}|\mathbf{x}, \theta) = \prod_{l=1}^{L} P(s_l|s_{<l}, \mathbf{x}; \theta)
\label{normalized-prob}
\end{equation}
where $s_{<l} \triangleq {s_1, s_2, .. , s_{l-1}}$ referring to generated tokens before the generation of $s_l$. \citet{kuhn2023semantic} simplifies the ensemble sampling in (\ref{posterior}) by using a single model in the ensemble due to the large size of foundation models. We follow the simplified version in the rest of the paper:
\begin{equation}
\begin{split}
P(\mathbf{s}|\mathbf{x}, D) \!\approx\! {P}(\mathbf{s}|\mathbf{x}, \theta) \!=\! \prod_{l=1}^{L} P(s_l|s_{<l}, \mathbf{x}; \theta).
\end{split}\label{new_def}
\end{equation}

\subsection{Length-Normalized Scoring}\label{sec:length_normalized_def}

One of the key issues with using sequence probability \( P(\mathbf{s}|\mathbf{x}, \theta) \) as a proxy for \( P(y|x, \theta) \) lies in its tendency to decrease as the sequence length increases. To overcome this issue, \citet{malinin2021uncertainty} uses length-normalized scoring function instead of sequence probability.\footnote{A scoring function $K$ takes two inputs: the predicted probability \( p \) of an event and its actual outcome \( o \), and returns a numerical score \cite{gneiting2007strictly}.} Length-normalized scoring $\tilde{P}(\mathbf{s}|\mathbf{x}, \theta)$ is defined as follows:
\begin{equation}
     \tilde{P}(\mathbf{s}|\mathbf{x}, \theta) = \prod_{l=1}^{L}  P(s_l|s_{<l}, \mathbf{x}; \theta)^{\frac{1}{L}},
\label{length-normalized-prob}
\end{equation}
which assigns equal weights to each token in the generation where these weights are inversely proportional to the sequence length $L$. Although length-normalized scoring $\tilde{P}(\mathbf{s}|\mathbf{x}, \theta)$ does not correspond to an actual probability distribution, \citet{malinin2021uncertainty} and \citet{kuhn2023semantic} consider $\tilde{P}(\mathbf{s}|\mathbf{x}, \theta)$ as auxiliary probabilities and replace the sequence probability ${P}(\mathbf{s}|\mathbf{x}, \theta)$ in (\ref{new_def}) with the length-normalized scoring given in (\ref{length-normalized-prob}).

\subsection{Entropy-Based UE for Generative LLMs}\label{entropy_section}

To obtain the entropy of the output for given input $\mathbf{x}$, \citet{malinin2021uncertainty} uses Monte-Carlo approximation over beam-sampled generations of a single model, as going through the entire answer set is infeasible due to its exponential computation complexity. Approximated entropy is defined as:
\begin{equation}
\mathcal{H}(\mathbf{x},\theta)\approx - \frac{1}{B} \sum_{b =1}^B \ln {\tilde{P}}(\mathbf{s}_b|\mathbf{x}, \theta),
\label{entropy}
\end{equation}
where $\mathbf{s}_b$ is an output sampled by beam-search and $B$ is the total number of sampled generations.

\citet{kuhn2023semantic} proposes an alternative entropy definition, named Semantic Entropy (SE), considering the meaning of the generations. They use the same entropy definition in (\ref{entropy}), but cluster sampled generations based on their meaning. For example, in response to the question ``What is the capital city of France?'', a model might output: ``Paris'' with score  \( \tilde{p}_1 \) and ``It's Paris'' with score \( \tilde{p}_2 \). 
While standard entropy in (\ref{entropy}) treats these as distinct outputs, SE clusters them together as they convey the same meaning in the question context, forming a single cluster \( \mathrm{c} \) with summed score \( \tilde{p}_1 + \tilde{p}_2 \).
More formally, cluster scoring is defined as:
\begin{equation}
 \tilde{P}(\mathrm{c} | \mathbf{x}, \theta) = \sum_{\mathbf{s}, \mathbf{x} \in \mathrm{c}} \tilde{P}(\mathbf{s}|\mathbf{x}, \theta).
\label{cluster-prob}
\end{equation}
SE follows from this cluster scoring $\tilde{P}(\mathrm{c} | \mathbf{x}, \theta)$:
\begin{equation}
SE(\mathbf{x},\theta)= - \frac{1}{|C|} \sum_{i=1}^{|C|} \log \tilde{P}(\mathrm{c}_i|\mathbf{x}, \theta),
\label{semantic-entropy}
\end{equation}
where $\mathrm{c}_i$ refers to each semantic cluster and $C$ is the set of all clusters. In Section \ref{theory}, we provide an alternative explanation for Semantic Entropy and length-normalized scoring from a classification task perspective.

Negative length-normalized scoring of the most probable answer, standard sequence entropy in (\ref{entropy}) and semantic entropy in (\ref{semantic-entropy}) are the most common probability-based UE methods for generative LLMs \cite{malinin2021uncertainty, kuhn2023semantic, chen2023quantifying, lin2023generating} and are visualized in Figure~\ref{fig:background}. All of these methods depend on length-normalized scoring which we aim to replace with our alternative scoring, MARS.

\section{Method}\label{method}

\subsection{Key Intuition}

Existing literature utilizes length-normalized scoring in UE as shown in (\ref{length-normalized-prob}), (\ref{entropy}), and (\ref{cluster-prob}).
Length-normalized scoring, given in (\ref{length-normalized-prob}), assigns equal importance/weight ($1/L$) to each token in the generated sentence. 
The normalization aims to compare the probabilities of short and long sequences more fairly
\cite{malinin2021uncertainty}. Such a normalization method may fall short in considering semantic contribution of tokens, even though it balances length differences across sequences. 

To illustrate, consider the following example: \textbf{Question:} ``Which planet is known as the Red Planet?'' \textbf{Generated Answer:} ``Mars is known as the Red Planet". In this answer, the word ``Mars'' is relatively more important as it directly addresses the question. Other words in the sentence primarily serve syntactic purposes or help achieve human-like answer. Thus, while designing a scoring function, we should give more importance/weight to the word ``Mars''. With this intuition, we want to replace length-normalized scoring and propose an alternative scoring function that assigns importance/weight to each word in the sentence considering both its contribution to the overall meaning in the given context and sequence length.

\subsection{Meaning-Aware Response Scoring}
Following our word importance intuition, we propose to replace length-normalized scoring $\tilde{P}(\mathbf{s}|\mathbf{x}, \theta)$ in (\ref{length-normalized-prob}), (\ref{entropy}), and (\ref{cluster-prob}) with Meaning-Aware Response Scoring (MARS). MARS is defined as:
\begin{equation}
    \bar{P}(\mathbf{s}|\mathbf{x}, \theta) = \prod_{l=1}^{L} P(s_l|s_{<l}, \mathbf{x}; \theta)^{w(\textbf{s},\mathbf{x}, L, l)},
\label{mars}
\end{equation}
where $w(\cdot)$ is the weighting function that assigns a weight to each token regarding the generated answer, question context, and sequence length.

We design $w(\cdot)$ as a convex combination of importance coefficient and $1/L$, which enables MARS to consider both sequence length and meaning contribution of tokens. Formally, we define

\begin{equation}
    w(\textbf{s},\mathbf{x}, L, l) \triangleq \frac{1}{2L} + \frac{u(\textbf{s},\mathbf{x},l)}{2},
\end{equation}
where $u(\cdot)$ is importance function taking three arguments: generated sequence $\mathbf{s}$, contextual information $\mathbf{x}$, and the position $l$ of a token within the sequence. The function $ u(\cdot) $ assigns an importance coefficient to each token, where this coefficient ranges between 0 and 1. Additionally, it ensures that the total sum of the importance coefficient for all tokens in a single generation $\mathbf{s}$ is 1. Next, we explain how to design the importance function $u(\cdot)$.


\subsection{Importance Function Design} \label{importance_func}
 
We design the token importance function $u(\cdot)$ 
by measuring the semantic impact of removing a specific token from the generated text. 
This evaluation of meaning is context-sensitive. In question-answer tasks, which is the focus of this work, the context is defined as the question itself.  
Thus, $u(\cdot)$ is  designed 
to determine the importance of each token based on its influence on the overall meaning of the response within the context of the question.

To measure the amount of semantic change in the given context, we employ a neural network model originally developed as a question-answer evaluator by \citet{bulian2022tomayto}. This model, called BERT matching (BEM), takes three inputs: question, ground truth answer, and predicted answer, returning a probability score  indicating answer correctness. For a question \( \mathbf{x} \) and a generated answer \( \mathbf{s} = \{s_1, s_2, \ldots, s_L\} \), we determine the importance of each token as follows: We mask token \( s_l \) in the generated answer and feed the question \( \mathbf{x} \), the original answer \( \mathbf{s} \), and masked response sequence \( \mathbf{s} \setminus \{s_l\} \) into the BEM model. The output \( o \), ranging from 0 to 1, indicates the impact of the masked token on answer correctness. A token \( s_l \) with substantial impact yields an output \( o \) close to 0, whereas a lesser impact results in an output closer to 1. Hence, we define \( 1 - o \) as the preliminary coefficient of \( s_l \). Once we compute preliminary coefficients for all tokens, we normalize them using a softmax function with a temperature parameter \( \tau \). In our experiments, we set \( \tau = 0.01 \).

\paragraph{Addressing Token Dependency.} 

Our initial approach for assigning importance coefficients to tokens assumes their semantic independence even though tokens often exhibit semantic inter-dependencies. For example, in the sentence ``Hamlet is written by William Shakespeare,'' tokens ``William'' and ``Shakespeare'' are intrinsically linked. Treating such tokens independently ignores linguistic nuances, so we refine our methodology. Instead of masking tokens individually, we mask tokens at the phrase level (details in Appendix~\ref{assign_phrase}). This approach acknowledges and preserves the inherent semantic relationships between closely related tokens, resulting in a more accurate and contextually aware assessment of token importance. In particular, a response $\textbf{s}=\{s_1, s_2, \ldots, s_L\}$ is composed of phrases $\{h_1, h_2, \ldots, h_K\}$, where each token $s_l$ belongs to a phrase $h_k$. We mask phrases one by one and find the importance coefficient of each phrase with BEM model. To translate phrase-level importance coefficients into token-level coefficients, we distribute the importance score to all tokens in the phrase equally. We summarize the enhanced algorithm in Appendix~\ref{algo-append}. Further, in Section~\ref{further_studies}, we show that allocating importance score only to the most uncertain token within a phrase also yields comparable results.


\begin{figure}[]
\begin{center}
\includegraphics[width=0.47\textwidth]{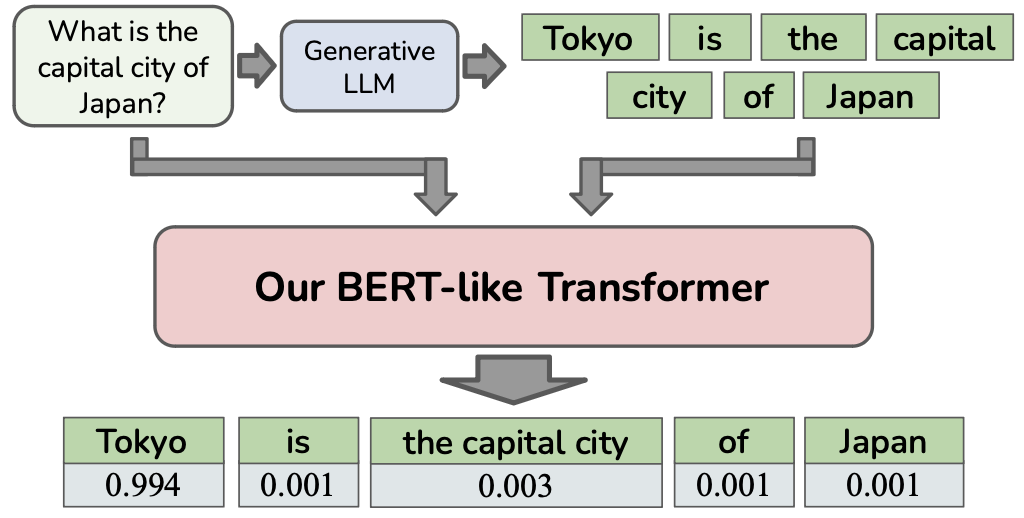}
\caption{Our Bert-like transformer model takes the question and the generated answer as inputs, and outputs phrases in the generated answer and corresponding importance coefficients. }
\label{fig:bert-model}
\end{center}
\vskip -0.3in
\end{figure}

    \paragraph{Reducing Computation.} The necessity of performing a separate neural network pass for each phrase to determine its importance score increases the computational load of the proposed approach. Additionally, detecting phrases themselves requires another neural network pass, further increasing the computational complexity. To address these challenges, we have developed a BERT-like neural network model with 110M parameters (a significantly smaller model compared to LLMs). This model is capable of performing both tasks simultaneously for a given sequence in a single neural network pass: it identifies phrases within the generated text and their importance scores (see Figure~\ref{fig:bert-model}). This dual-functionality substantially reduces the computational cost, making the algorithm more efficient and scalable. For detailed model architecture and performance metrics, please refer to Appendix~\ref{bert_model}. 


\section{Understanding Generative LLM Probabilities from a Classification Perspective} \label{theory} 
In classification tasks, the \textit{class probability} reflects the model's confidence in assigning a specific class to an input. It is inherently tied to the \textit{semantics} of the class. For instance, if a well-calibrated classifier gives a 75\% probability to the label ``cat'' for a given image, it suggests a 75\% likelihood that the given image is indeed a cat. This output probability is not only a numerical value; it conveys a semantic understanding of the image content as a cat. To understand previous works \cite{kuhn2023semantic, malinin2021uncertainty}  from a classification perspective, we propose a new random variable that is directly related to the semantics of the output.

Let $Y$ be a random variable with arbitrary dimension corresponding to the meaning of the sequences generated by an LLM parametrized with $\theta$. The values of $Y$ can be the set of all possible meanings of generated sequences and their contexts. Formally, the set is $\{g(\textbf{s},\textbf{x})\}_{\textbf{s} \in \mathcal{S}, \textbf{x} \in \mathcal{X}}$, where $g(\cdot)$ is the meaning function that takes generated sentence $\textbf{s}$ and context $\textbf{x}$ as inputs and returns the meaning as output. A well-calibrated probability distribution of $Y$ ensures that if a model's generation is more likely to be correct than another, the corresponding probability should be higher. Mathematically, $P(Y =  g(s_1,x_1) | \theta)$ > $P(Y =  g(s_2,x_2) | \theta)$ if the model's response $s_1$ to question $x_1$ is more likely to be correct than response $s_2$ to question $x_2$. This is indeed similar to the classification example where well-calibrated class probabilities reflect the output's actual probability of correctness \cite{guo2017calibration}. Previous works and MARS have heuristic design choices to make $Y$ more calibrated to get a better uncertainty estimation.


\citet{malinin2021uncertainty} considers  $g(\cdot)$ as a one-to-one function which means that each unique sentence in the given context corresponds to different meanings. To achieve better calibration of $Y$ they define its distribution using length-normalized scoring of the generated sequences. This approach ensures a fair comparison between long and short sequences, thereby improving the calibration of $Y$. More formally
\begin{equation}
\begin{split}
    P(Y =  y | \theta) = \frac{\tilde{P}(\mathbf{s}|\textbf{x}, \theta)}{\sum_{\textbf{s} \in \mathcal{S}, \textbf{x} \in \mathcal{X}}\tilde{P}(\mathbf{s}|\textbf{x}, \theta)},
\end{split}\label{length-normalized}
\end{equation}
where $y = g (\textbf{s},\textbf{x})$ and  $\tilde{P}(\mathbf{s}|\textbf{x}, \theta)$ is the length-normalized scoring  defined as $\prod_{l=1}^{L} P(s_l|s_{<l}, x; \theta)^{1/L}$. To make the distribution of $Y$ a valid probability distribution, we normalize each $\tilde{P}(\mathbf{s}|\textbf{x}, \theta)$ by the sum of all possible scores, making their summation 1. By defining $Y$ as above, we essentially create an actual probability distribution of length-normalized scoring.


On the other hand, \citet{kuhn2023semantic} claims different sequences can have equal meaning. By considering $g(\cdot)$ as a many-to-one function,
we can define the distribution of Y with their proposal as follows:
\begin{equation}
\begin{split}
    P(Y = y  | \theta)
    &=  \frac{\sum_{\textbf{s},\textbf{x} \in c_y} \tilde{P}(\mathbf{s}|\textbf{x}, \theta)}
    {\sum_{\textbf{s} \in \mathcal{S}, \textbf{x} \in \mathcal{X}}\tilde{P}(\mathbf{s}|\textbf{x}, \theta)}
    \label{cluster}
\end{split}
\end{equation}
where $c_y$ corresponds to the meaning cluster, formally written as $c_y = \{\textbf{s},\textbf{x} | g(\textbf{s},\textbf{x}) = y\}$. By employing this new probability definition within the standard entropy calculation in (\ref{entropy}), we obtain the concept of semantic entropy as follows
\begin{equation}
\begin{split}
    SE(\mathbf{x},\theta)= - \frac{1}{B} \sum_{b=1}^{B} \log P(Y = y_b  | \theta)
\end{split}
\end{equation}
With the new random variable $Y$, we essentially write the semantic entropy as the standard Monte-Carlo approximated entropy over a total of $B$ distinct meanings.

Notice that the normalization term $\sum_{\textbf{s} \in \mathcal{S}, \textbf{x} \in \mathcal{X}}\tilde{P}(\mathbf{s}|\textbf{x}, \theta)$ featured in both (\ref{length-normalized}) and (\ref{cluster}), acts as a constant across all \( P(Y = y | \theta) \) calculations, ensuring that \( Y \) conforms to a valid probability distribution. Therefore, it only shifts the proposed UE scores which does not affect the performance of accurately predicting the correctness of the model generation. Moreover, by introducing the random variable \( Y \), we not only provide an alternative explanation of the previous works but also create flexibility to define new distributions for $Y$ which may potentially have better calibration and improve the existing UE tools. 

Using the definition of $Y$, we can also rationalize our scoring function MARS.  
We replace the length-normalized scoring function with MARS as in (\ref{mars}). MARS considers the semantic contribution of tokens, assigning higher weights to those critical for the correctness of the answer. This emphasis on key tokens potentially makes MARS a more effective scoring function for achieving better calibration of $Y$.



\section{Experiments}
\subsection{Experimental Design}
In the UE context, we expect that if the model is uncertain about the generated answer, then the answer should be less reliable and tend to be incorrect.

\paragraph{Datasets.} We use three closed-book Question-Answer (QA) datasets for evaluation: TriviaQA \cite{joshi2017triviaqa}, Natural Questions \cite{kwiatkowski2019naturalqa}, and WebQA \cite{WebQA21}.  We give 
further details in Appendix~\ref{exp_details}.

\paragraph{Models.} Our evaluation consists of 5 popular open-source LLMs. First two models are Llama-7B and Llama-7B-chat, where the latter one is fine-tuned for dialogue use cases \cite{touvron2023llama2}. We also use Mistral-7B \cite{jiang2023mistral} as well as Falcon-7B \cite{falcon40b} which is fine-tuned on a mixture of chat/instruct datasets. To extend our analysis to larger models, we include Llama-13B \cite{touvron2023llama2}. We do not perform any further training on these models, rather we use their pre-existing configurations. Following \citet{kuhn2023semantic}, we abstain from assuming any ensemble of the models, considering the significant size and time requirements associated with LLMs.

\paragraph{Baselines.} As we focus on the probability-based UE methods, we do not include heuristic-based and black-box methods. We use 3 SOTA probability-based UE methods as baselines (see Figure~\ref{fig:background} for visualization): \textbf{1.} Negative length-normalized score (Confidence), which provides the confidence score of the most likely generation only by using its token probabilities as in (\ref{length-normalized-prob}). \textbf{2.} Entropy as in (\ref{entropy}), which requires generating multiple answers to obtain the score for the most likely answer. \textbf{3.} Semantic Entropy (SE), which considers the meaning of the generated answer while computing entropy, as shown in (\ref{semantic-entropy}). 
All 3 baselines depend on length-normalized scoring. We replace length-normalized scoring with MARS and arrive at Confidence + MARS, Entropy + MARS, SE + MARS.

\paragraph{Metrics.} Following previous works \cite{malinin2021uncertainty, kuhn2023semantic}, we use Area Under the Receiver Operating Characteristic Curve (AUROC) score for our UE performance metric. {AUROC quantifies a method's ability to distinguish between two classes by plotting the true positive rate against the false positive rate for various threshold values. AUROC score is the area under this curve, ranging from 0 to 1.} Higher AUROC score indicates a superior performance, while a score of 0.5 implies a random chance. In our case, ground truth is the correctness\footnote{We use GPT-3.5-turbo for evaluating the correctness of the model, as in \cite{lin2023generating, chen2023quantifying}.} of the model response to the question and the prediction is the output of an UE method. 



\begin{table*}[!htbp]
\centering
\fontsize{9.3}{8.2}\selectfont
\begin{tabular}{l|p{0.08cm}l ccccc}
\toprule
 && Method & \textbf{Llama2-7b} & \textbf{Llama2-7b-chat} & \textbf{Mistral-7b}& \textbf{Falcon-7b}& \textbf{Llama2-13b} \\
\midrule[\heavyrulewidth]

\multirow{6}{*}{\rotatebox{90}{\textbf{TriviaQA}}}
&&Confidence    & 70.18 & 70.40 & 72.55 & 68.47 & 68.19\\
&&Entropy       & 69.70 & 69.94 & 72.57 & 69.10 & 69.04\\
&&SE            & 81.10 & 76.19 & 82.17 & 76.78 & 79.49\\
\cmidrule{2-8}
&\multirow{3}{*}{\rotatebox{90}{\textit{Ours}}}
&Confidence + MARS & \textbf{75.06} & \textbf{74.23} & \textbf{77.97} & \textbf{72.95} & \textbf{73.99} \\
&&Entropy + MARS &\textbf{75.94} & \textbf{73.82} & \textbf{78.51} & \textbf{72.87} & \textbf{74.95}\\
&&SE + MARS      & \textbf{82.22} & \textbf{77.67} & \textbf{83.63} & \textbf{77.48} & \textbf{81.00}\\
\midrule[\heavyrulewidth]

\multirow{6}{*}{\rotatebox{90}{\textbf{NaturalQA}}}
&&Confidence         & 68.56 & 65.98 & 69.54 & 63.78 & 68.56\\
&&Entropy      & 67.08 & 65.23 & 68.05 & 63.28 & 68.34\\
&&SE           & 72.47 & 68.66 & 75.12 & 70.41 & 73.56\\
\cmidrule{2-8}
&\multirow{3}{*}{\rotatebox{90}{\textit{Ours}}}
&Confidence+ MARS     & \textbf{69.81} &\textbf{67.86} & \textbf{71.36} & \textbf{68.30} & \textbf{70.88}\\
&&Entropy + MARS &  \textbf{69.32} & \textbf{67.41} & \textbf{70.71} & \textbf{67.51} & \textbf{70.63}\\
&&SE + MARS      &  \textbf{72.75} & \textbf{69.43} & \textbf{75.50} & \textbf{71.24} & \textbf{73.89}\\
\midrule[\heavyrulewidth]

\multirow{6}{*}{\rotatebox{90}{\textbf{WebQA}}}
&&Confidence         & 64.76 & 64.06 & 65.66 & 66.56 & 62.60\\
&&Entropy      & 64.04 & 63.82 & 64.15 & 65.98 & 62.11\\
&&SE           & 69.44 & 67.11 & 69.51 & 73.16 & 67.31\\
\cmidrule{2-8}
&\multirow{3}{*}{\rotatebox{90}{\textit{Ours}}}
&Confidence + MARS     & \textbf{66.04} & \textbf{64.48} & \textbf{67.16} & \textbf{68.26} & \textbf{64.23}\\
&&Entropy + MARS &\textbf{65.83} & \textbf{64.69} & \textbf{65.76} & \textbf{68.44} & \textbf{64.02}\\
&&SE + MARS      & \textbf{69.88} & \textbf{67.27} & \textbf{69.86} & \textbf{73.57} &\textbf{67.75}\\
\bottomrule
\end{tabular}
\caption{AUROC performance of UE methods in various datasets with different pre-trained LLMs.}
\vskip -0.1in
\label{tab:main_results}
\end{table*}

\subsection{Main Results}\label{results}
We present our detailed results in Table~\ref{tab:main_results}. Upon closer examination of the results, it becomes apparent that the application of MARS consistently improves all baseline methods across various datasets and models. Specifically, MARS yields improvements of up to 5.8 points for Confidence, 6.24 points  for Entropy, and 1.51 points for SE.

It is crucial to mention that the choice among the baselines depends on the available computational resources. Confidence score is the least resource-intensive, requiring only a single output generation. Entropy, on the other hand, demands multiple generations (set to 5 in our experiments). SE is the most computationally demanding, needing both multiple generations and \( O(n^2) \) Natural Language Inference (NLI) model passes for clustering, where \( n \) represents the number of generations.

One of the main contributions of MARS becomes evident when we compare SE with Confidence+MARS or Entropy+MARS. With our method, we are able to increase the scores of Confidence+MARS and Entropy+MARS to a level they can compete with basic SE. Consequently, given the computational overhead of SE, Confidence+MARS and Entropy+MARS emerge as more practical and desirable alternatives. Furthermore, in scenarios where sampling (i.e., multiple answer generation) is not feasible, the improvement offered by MARS to Confidence method becomes crucial with an average increase of 2.8 points. We note that the additional computational and memory demands of MARS are relatively minor, approximately 1.5\% of the 7b models and 0.8\% of the 13b models, because MARS's importance function is implemented with 110M Bert-like model.

\begin{table}[!htbp]

\centering
\fontsize{8.5}{8}\selectfont
\begin{tabular}{p{0.08cm}l|cc}
\toprule
 & Method & \textbf{Llama2-7b} & \textbf{Mistral-7b} \\
\midrule[\heavyrulewidth]
\multirow{3}{*}{\rotatebox{90}{\textit{Token}}}
&Confidence + MARS     & 72.53 & 75.31\\
&Entropy + MARS        & 74.46 & 77.58\\
&SE + MARS             & 81.55 & 83.25\\
\midrule[\heavyrulewidth]
\multirow{3}{*}{\rotatebox{90}{\textit{Phrase}}}
&Confidence + MARS     & \textbf{75.06} & \textbf{77.97}\\
&Entropy + MARS        & \textbf{75.94} & \textbf{78.51}\\
&SE + MARS             & \textbf{82.22} & \textbf{83.63}\\
\bottomrule

\end{tabular}
\caption{AUROC score of UE methods + MARS with token/phrase-level importance functions on TriviaQA. }
\label{tab:word_phrase_results}
\end{table}

\subsection{Ablation Studies} \label{further_studies}

\paragraph{Effect of Phrase Separation.}
In Section \ref{importance_func}, we suggest using a phrase-level separation instead of token-level separation in designing the importance function so that tokens having strong relations are evaluated together on their semantic impact on the sequence. To validate this design, we conduct an experiment where we revert to token-level separation. The results in Table~\ref{tab:word_phrase_results} demonstrate that while token-level separation outperforms other baselines, phrase-level separation consistently yields superior results, reaffirming the efficacy of our approach.

\begin{table}[!t]
\centering
\fontsize{8.5}{8}\selectfont
\begin{tabular}{c|l| cccc}
\toprule
 Method & Distribution & \textbf{Llama2-7b} & \textbf{Mistral-7b} \\
\midrule[\heavyrulewidth]

\multirow{3}{*}{{\thead{Confidence\\+ MARS}}}
&Min     & 69.92 & 72.20\\
&Max     & \textbf{75.13} & 77.73 \\
&Equal     & 75.06 & \textbf{77.97}\\
\midrule[\heavyrulewidth]
\multirow{3}{*}{{\thead{Entropy\\+ MARS }}}
&Min        & 70.56 & 72.75\\
&Max        & \textbf{77.11} & \textbf{79.22} \\
&Equal        & 75.94 & 78.51\\
\midrule[\heavyrulewidth]
\multirow{3}{*}{{\thead{SE +\\MARS}}}
&Min             & 81.67 & 82.33\\
&Max             & 82.07 & \textbf{83.62} \\
&Equal             & \textbf{82.22} & \textbf{83.63}\\
\bottomrule
\end{tabular}
\caption{AUROC score of UE methods + MARS with different coefficient distributions in phrases in importance function on TriviaQA. }
\vspace{-4mm}
\label{tab:min_max_results}
\end{table}

\paragraph{Importance Coefficient Distribution in Phrases.}
In Section \ref{importance_func}, we state that we equally distribute the importance of phrases to each token.  Alternative distribution strategies might include prioritization of the least or most uncertain token. Those strategies assign the phrase importance coefficient to the least or most uncertain token of that phrase. In Table~\ref{tab:min_max_results}, we provide AUROC performances when different distribution strategies are adopted.   Notably, we find that max-uncertain distribution is nearly as effective as our adopted equally assigning approach. In contrast, the min-uncertain assigning strategy underperforms. This outcome can be contextualized with a hypothetical scenario: Consider the model's response is ``Shakespeare'' to the query ``Who wrote Hamlet?'', which is tokenized into ``Shake'' and ``-speare''. Once ``Shake'' is produced, the subsequent arrival of ``-speare'' is almost assured. The uncertainty primarily resides in the token ``Shake'', making the probability of ``-speare'' relatively uninformative. Consequently, focusing on the least uncertain (most uninformative) token in a phrase drops the performance of MARS significantly, and focusing on the most uncertain token only is still reasonable.

\begin{figure*}[!htbp]
\begin{center}
\includegraphics[width=\textwidth]{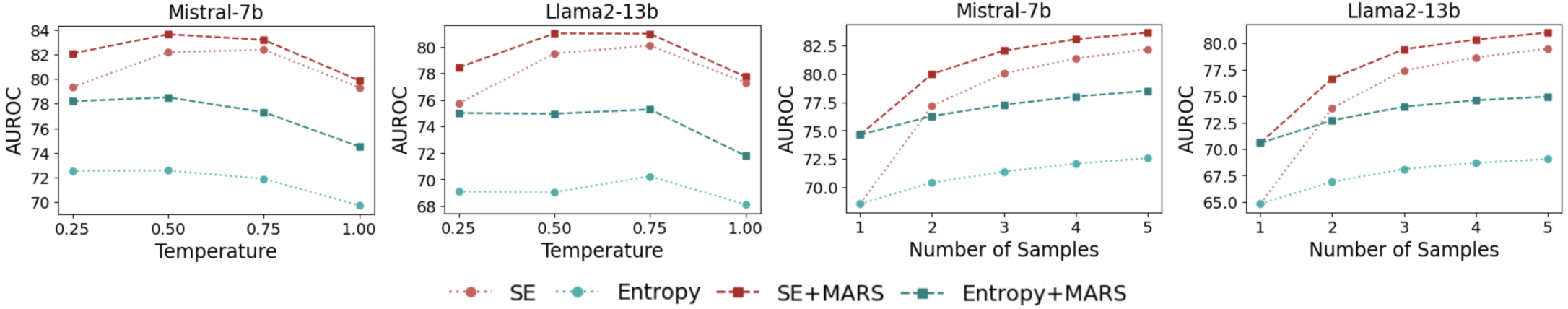}
\caption{AUROC scores for various temperatures and sampling numbers.}
\label{fig:temp_samp}
\end{center}
\vskip -0.25in
\end{figure*}

\subsection{Effect of Sampling Hyperparameters}
We explore the influence of key hyperparameters on the performance of UE methods that rely on sampling, specifically Entropy and SE. We focus on two critical hyperparameters: Temperature, which adjusts the diversity of the sampling process, and the number of sampling, which dictates how many samples are sampled in entropy calculation.
\paragraph{Temperature.} The temperature parameter determines the smoothness of the probabilities while sampling. A higher (lower) temperature value indicates more (less) diverse sampling. Figure \ref{fig:temp_samp} presents the AUROC scores for Entropy, SE, and their enhancements via MARS for the Llama2-13b and Mistral-7b models on the TriviaQA dataset. The improvement of MARS is consistent for all temperature values. The choice of temperature is application-dependent: higher temperatures are advisable for tasks demanding creativity, whereas lower temperatures are preferable for applications where consistency is important.


\paragraph{Number of Sampling.} The number of sampled sequences is important for entropy and semantic entropy calculation. More sampling leads to better entropy estimation; however, the cost also increases. Beyond the sampling expense, SE incurs an additional cost from Natural Language Inference (NLI) model passes, a point elaborated in Section \ref{results}. In Figure \ref{fig:temp_samp}, we provide the AUROC performance of Llama2-13b and Mistral-7b models on TriviaQA with various sampling numbers. Notably, the efficacy of MARS remains stable across diverse sampling numbers, with its advantages becoming more obvious under lower sampling numbers.


\subsection{UE in Medical QA Dataset} \label{medical_section}
Next, we evaluate the UE methods using a medical QA dataset. Publicly available medical QA datasets typically fall into two categories: those with multiple-choice questions \cite{pmlr-v174-pal22a, kotonya-toni-2020-explainable, jin2021disease} and  
those without clear ground truths 
\cite{Zhu2019AHA,zhu-etal-2020-question}. To tackle this, we create a subset from the MedMCQA multiple-choice dataset \cite{pmlr-v174-pal22a},
selecting questions that can be answered objectively without multiple choices. 
For this, we collaborate with medical professionals to ensure the accuracy and relevance of the selected questions, yielding a dataset of 415 samples. We use AdaptLLM's Medicine-Chat \cite{cheng2023adapting}, a medical-domain adapted LLaMA-2-Chat-7B model\footnote{https://huggingface.co/AdaptLLM/medicine-chat}. To evaluate the correctness of model-generated responses, 
we leverage GPT-4 \cite{openai2023gpt4} and assess response validity in the medical domain.

In Table \ref{tab:medical_res}, we provide the AUROC performance of the UE methods. Although MARS still consistently improves the performance of probability-based UE methods, AUROC scores are still low compared to Table~\ref{tab:main_results}. This might be because of the nature of medical questions. General knowledge questions mostly require a straight, single-sentence answer. On the other hand, although we curated closed-ended questions, medical questions still require a more complex explanation spanning multiple sentences. This difference between domains can affect the prediction performance of the probability-based methods. This observation emphasizes the necessity for further investigation across various specialized fields, including medicine and law. Customized explorations are essential to address domain-specific challenges and optimize UE methods accordingly.

\begin{table}[!htbp]

\centering
\fontsize{8.7}{8}\selectfont
\begin{tabular}{p{0.08cm}l|c}
\toprule
 & Method & \textbf{Medicine-Chat-7b}\\
\midrule[\heavyrulewidth]
\multirow{3}{*}{}
&Confidence     & 62.41 \\
&Entropy        & 59.58 \\
&SE             & 62.89 \\
\midrule[\heavyrulewidth]
\multirow{3}{*}{\rotatebox{90}{\textit{Ours}}}
&Confidence + MARS     & \textbf{62.89}\\
&Entropy + MARS        & \textbf{60.33}\\
&SE + MARS             & \textbf{64.48}\\
\bottomrule
\end{tabular}
\caption{AUROC score of UE methods on medical QA. }
\label{tab:medical_res}
\vskip -0.2in
\end{table}

\section{Conclusion}
We introduce Meaning-Aware Response Scoring (MARS), a novel scoring function designed to replace length-normalized scoring in probability-based UE methods when evaluating generative LLMs. MARS consistently and significantly boosts the performance of current probability-based UE methods with minimal additional computational overhead. The efficacy of MARS is shown in three closed-book and closed-ended question-answer datasets and a medical question-answer dataset.   


\section{Limitations}
The importance function model within MARS utilizes an unsupervised methodology, leveraging pre-existing models for its formulation. Nonetheless, the performance of MARS can potentially be further enhanced by using human labelers to assign importance coefficients for training the importance function model. Besides, our analysis is limited to the closed-ended question-answering domain in English, where a question has an objective ground-truth answer(s). Extensive analysis of MARS and other probability-based UE methods on open-ended question-answering tasks and other languages are beyond the scope of the current study and are left as future work.

\section{Ethics Statement}
Although probability-based UE methods combined with MARS have a remarkable prediction performance on the correctness of generative LLM outputs, it is crucial to acknowledge that these methods do not achieve 100\% accuracy. Besides, as LLMs may have biases against gender, ethnicity, age, etc., probability-based methods can carry those biases to UE outputs. Thus, one should be aware of these potential risk factors before employing such probabilistic UE methods in real-world systems. Ensuring fairness, transparency, and accountability in the deployment of these technologies is important in mitigating risks and fostering trust in their application.

\section{Acknowledgement}
This work is supported in part by a research gift from USC-Amazon Center on Secure and Trusted Machine Learning\footnote{https://trustedai.usc.edu} . Also, we are thankful to MD. Mehmet Sahin Ulukanli and MD. Burak Yaldiz for their invaluable collaboration and expertise in the development of the Medical QA dataset.

\bibliography{custom}
\bibliographystyle{acl_natbib}

\appendix

\section{Related Works}\label{related-work}
Uncertainty Estimation (UE) has emerged as a vital concept in various machine learning domains, particularly in Natural Language Processing (NLP). The study of \citet{xiao2022uncertainty} concentrates on the UE for tasks like common-sense reasoning and sentiment analysis; \citet{jiang-etal-2021-know} explores model calibration for UE in the context of multiple-choice question answering; \citet{desai2020calibration} tackles the challenge of UE in specific NLP tasks such as paraphrase detection and natural language inference. These studies represent just a fraction of the UE works in the field of NLP and there is an expanding corpus of research focusing on the investigation of UE in NLP \cite{hu2023uncertainty, xiao2018quantifying, vazhentsev2022uncertainty}. The vast majority of these studies only focus on classification and regression tasks, unlike our work where the goal is to study UE for generative LLMs. 


Few recent works deal with UE of generative LLMs. \citet{xiao2020wat} and \citet{fomicheva2020unsupervised} propose heuristic-based uncertainty metrics for generative LLMs considering machine translation. \citet{chen2023quantifying}, \citet{lin2023generating}, \citet{cohen-etal-2023-lm}, and \citet{kadavath2022language} propose black-box UE methods for generative LLMs under the assumption that the token probabilities are not accessible. Although these works have experimental validation, they lack a mathematical foundation. \citet{malinin2021uncertainty} is the first study adapting popular uncertainty tools in Bayesian UE literature to the generative LLMs. The main idea of \citet{malinin2021uncertainty} is to utilize length-normalized scoring in computing the entropy of the LLM answers. 
A more recent approach by \citet{kuhn2023semantic} further improves this result by introducing the concept of semantic entropy, which considers the meaning of the generated sentences in entropy calculation in uncertainty prediction. Our work is distinct from these works as we no longer utilize length-normalized scoring. Instead, we utilize the proposed MARS in entropy computations, by also taking into consideration token importance to the answer correctness , thereby achieving an improved UE performance.

\subsection{Discussion of the Differences with TokenSAR}
There is a recent work that also considers the meaning of the words in the generation to estimate uncertainty \cite{duan2023shifting}. The fundamental difference with our work is that \cite{duan2023shifting}'s method is designed as an alternative to the existing probability-based uncertainty methods, whereas, in our work, we propose a scoring function, i.e., MARS, which is compatible with all existing probability-based uncertainty estimation methods. This implies that one can in fact utilize MARS within the framework of \cite{duan2023shifting}. In particular, in \cite{duan2023shifting}, authors propose three schemes: TokenSAR (token-level weight assignment), SentSAR (sentence-level weight assignment), and SAR (both token and sentence-level weight assignment). SentSAR and SAR are orthogonal to MARS. SAR is the version of SentSAR where the scoring function in SentSAR is replaced with TokenSAR. In a similar fashion, MARS can be incorporated into the SentSAR approach instead of the TokenSAR.

Thus, we need to discuss our distinction from TokenSAR, which can also be considered as a scoring function. To avoid confusion and clarify our unique approach, below we discuss our distinction from TokenSAR. 

\begin{itemize}
    \item \textbf{MARS uses BERT-Matching instead of sentence similarity}: In our algorithm, we utilize the BERT-Matching (BEM) model which takes the question, ground truth answer, and the generated answer as inputs and returns the probability of the generated answer being correct. To assign importance weights, we remove a set of tokens (a phrase) from the generated sentence and pass the question, generated answer (as ground truth) and token-removed generated answer to the BEM model. We set (1 - output) as the importance weight and normalize weights at the end. We further improve this process by fine-tuning a BERT-like model and increase efficiency (explained in the third bullet). TokenSAR uses sentence similarity model (cross-encoder Roberta-Large) unlike our approach. Sentence similarity model takes two input sentences to measure similarity and they concatenate the question for both inputs. However, we argue that using the BEM model achieves better performance since the goal is to find a token’s importance based on its contribution to the correctness of the generated answer. This difference becomes more visible when the answer is longer and more complex as we demonstrate in the below example. In particular, TokenSAR fails to detect words that actually answer the question so that it (almost) returns uniform importance values. On the other hand, MARS successfully finds the important words and assigns higher weights to them. Let’s consider the following example:

Question: What is the tallest building in the world?

Generated Answer: The Burj Khalifa in Dubai, soaring into the sky, holds the distinction of being the tallest building in the world, a marvel of modern engineering and architecture.

To this question-answer pair, MARS returns the following importance weight assignment:

\texttt{The Burj Khalifa (0.8428) in (0.0082) Dubai (0.0083) , (0.0082) soaring (0.0084) into (0.0082) the sky (0.0082) , (0.0082) holds (0.0083) the distinction (0.0083) of (0.0082) being (0.0082) the tallest building (0.0082) in (0.0082) the world (0.0082) , (0.0082) a marvel (0.0083) of (0.0083) modern engineering and architecture (0.0088) . (0.0083) }

On the other hand, to the same pair, TokenSAR returns the following importance weight assignment:

\texttt{The (0.0225) Bur (0.0228) j (0.0318) K (0.0228) hal (0.0228) ifa (0.0319) in (0.0227) Dub (0.0253) ai (0.0232) , (0.0228) so (0.0237) aring (0.0294) into (0.0228) the (0.0228) sky (0.0235) , (0.0229) holds (0.0228) the (0.0227) distinction (0.0234) of (0.0228) being (0.0228) the (0.0229) tall (0.0235) est (0.0227) building (0.0228) in (0.0228) the (0.0228) world (0.0230) , (0.0229) a (0.0228) mar (0.0232) vel (0.0232) of (0.0230) modern (0.0540) engineering (0.0725) and (0.0336) architecture (0.0328) . (0.0232)}

In this example, although the phrase “The Burj Khalifa” is the key word answering the question, TokenSAR assigns low weights to its tokens. In fact, according to TokenSAR, tokens of the phrase “The Burj Khalifa” are as important as some of the words/phrases that appear in the question itself such as “the tallest building”. This is not ideal as TokenSAR cannot distinguish between the actual answer and filler words. However, our proposed MARS is able to actually find the important words in the answer thanks to the BEM model we employ during weight assignment.

\item \textbf{MARS addresses token dependencies and process phrases instead of tokens}: As we explain in Section \ref{method}, we first divide a generated answer into phrases and then assign scores to each of those phrases by using the procedure described in the first bullet point. On the other hand, \cite{duan2023shifting} assumes that each generated token is meaningly independent so that they remove tokens from the generation one-by-one and assign importance scores accordingly. However, as we show in Table \ref{tab:word_phrase_results}, ignoring token dependencies negatively affects the performance of uncertainty methods. In this sense, our MARS provides a more careful importance score assignment (as we demonstrate in the above example).

\item \textbf{MARS is computationally efficient at inference}: As we mention in Section \ref{method}, we improve the computational performance of MARS by fine-tuning a BERT-like model that gives importance scores with phrases in a single forward pass (this is an improvement over the algorithm described in the first bullet). That is, our importance assignment does not depend on the number of tokens in the generated sentence. In stark contrast, \cite{duan2023shifting} uses cross-encoder Roberta-Large and their algorithm requires a number of tokens times forward pass for a single generated sentence. Moreover, cross-encoder Roberta-Large has approximately 355M parameters. However, we run our 110M BERT-like model only one time for the generated sentence no matter its length. That is why for a generation comprised of 10 tokens, MARS is ~30x computationally more efficient than \cite{duan2023shifting}.
\end{itemize}

\begin{table*}[!htbp]
\centering
\resizebox{\textwidth}{!}{%
\begin{tabular}{p{4cm}|p{5cm}|c}
\toprule
\multicolumn{1}{c|}{ \textbf{Question}} &\multicolumn{1}{c|}{ \textbf{Answer}} &\multicolumn{1}{c}{ \textbf{Output}}  \\
\midrule[\heavyrulewidth]

\multicolumn{1}{m{4cm}|}{Which planet is known as Red Planet?}&
\multicolumn{1}{m{4cm}|}{It is Mars }&
\begin{tabular}{ccccc}
        It & is & Mars  \\ 
    0.017 & 0.017  & 0.956   \\
\end{tabular} \\

\midrule[\heavyrulewidth]
\multicolumn{1}{m{4cm}|}{What is the capital city of Japan?}&
\multicolumn{1}{m{4cm}|}{Tokyo is the capital city of Japan} &
\begin{tabular}{ccccc}
        Tokyo & is & the capital city  &of & Japan  \\ 
    0.994 & 0.001  & 0.003  & 0.001 & 0.001  \\
\end{tabular} \\

\midrule[\heavyrulewidth]
\multicolumn{1}{m{4cm}|}{Which element has the chemical symbol "O"?}&
\multicolumn{1}{m{4cm}|}{The chemical symbol "O" represents Oxygen} &
\begin{tabular}{cccc}
    The chemical symbol & "O" & represents  &Oxygen \\ 
    0.01 & 0.01  & 0.003  & 0.976 \\
\end{tabular} \\



\bottomrule
\end{tabular}}
\caption{Sample outputs of our BERT-like model used for importance function. Question and answer are given to the model as input, and the model divides the answer into phrases while assigning importance score. }
\label{tab:output_samples}
\end{table*}

\section{Training of BERT-like Model for Importance Function} \label{bert_model}

As described in Section \ref{importance_func}, we optimize the computational efficiency of MARS by training a single Bert-like model with 110M parameters to execute the importance function. This model is an adaptation of the pre-trained Bert-base-uncased\footnote{https://huggingface.co/bert-base-uncased}, modified by removing its last layer and incorporating two independent fully-connected (FC) layers. The first FC layer focuses on phrase detection with two output logits: ``Begin Phrase'' (BP) and ``Inside Phrase'' (IP), and classifies each token as BP if it marks the start of a phrase or as IP otherwise. This setup enables sentence segmentation into phrases. The second FC layer, tasked with assigning importance coefficients, produces a single output logit for each token's importance coefficient. 

For training data, we take a subset of 69192 question samples from the TriviaQA training set and questions of the whole training set of NaturalQA consisting of 87925. Then, we use these questions as input and feed them to all 7B-sized baseline models (Llama2-7b, Llama2-7b-chat, Mistral-7b, Falcon-7b) to yield the responses. This provides us with question-answer pairs. We use the Flair phrase chunking model to determine phrase labels in the answers, as described in Appendix~\ref{assign_phrase}. For importance coefficient labels per token in the responses, we follow Algorithm ~\ref{algo}. 

Sample outputs of our model are provided in Table~\ref{tab:output_samples}.  Here, question and answer are inputs to the model, and the model divides the answer into phrases while assigning importance score to them.

We train the model only for 1 epoch with 5e-5 learning rate and 32 batch size. The training process involves a convex combination of two loss functions: cross-entropy for phrase chunking and negative log-likelihood for importance coefficient assignment, with equal weight assigned to both losses. Table \ref{tab:bert_loss} displays the training and validation losses at the end of the training, indicating that our training objectives are effectively generalizable to test sets.
\begin{table}[!htbp]

\centering
\begin{tabular}{l|cc}
\toprule
& \textbf{Classification} &\textbf{Scoring} \\
& \textbf{Loss} &\textbf{Loss} \\
\midrule[\heavyrulewidth]

Train      & 0.0275 & 0.1957 \\
Validation & 0.0205 & 0.1901\\
\bottomrule

\end{tabular}
\caption{Train and validation loss values calculated at the end of training of BERT-like importance model. Classification loss stands for cross-entropy loss for phrase chunking, and Scoring loss indicated negative log-likelihood loss for importance coefficient.}
\label{tab:bert_loss}
\end{table}

\subsection{Dividing a Sentence to Phrases} \label{assign_phrase}
To divide a sentence into phrases, we use the Flair phrase chunking model\footnote{https://huggingface.co/flair/chunk-english} \cite{akbik2018coling}, that uses 10 tags which are adjectival, adverbial, conjunction, interjection, list marker, noun phrase, prepositional, particle, subordinate clause and verb phrase. For example, the Flair model divides the sentence ``The happy man has been eating at the dinner'' as ``The happy man'', ``has been eating'', ``at'', ``the diner''. 

\subsection{Pseudocode of the Importance Function Algorithm} \label{algo-append}

The pseudocode of the importance function algorithm is given in Algorithm \ref{algo}.
\begin{algorithm}[!htbp]
\caption{Phrase-Level Importance Function}
\begin{algorithmic}[1]
\Statex \textbf{Input:} Question $\mathbf{x}$, generated answer $\mathbf{s} = \{s_1, s_2, \ldots, s_L\}$, phrases $\{h_1, h_2, \ldots, h_K\}$, token probabilities $\{p_i = P(s_i|s_{<i}, \mathbf{x}; \theta) \}_{s_i \in \mathbf{s}}$, temperature $\tau$
\Statex \textbf{Output:}  Importance scores $I$
\Statex $I \gets$ []
\For{$k = 1$ to $K$}
    \State  $\mathbf{s}_{\text{masked}} \gets \mathbf{s} \setminus \{s_l\}_{s_l \in h_k}$
    \State $o_k \gets BEM(\mathbf{x}, \mathbf{s}, \mathbf{s}_{\text{masked}})$
    \For{each token $s_l$ in phrase $h_k$}
        \State $I[l] \gets (1 - o_k) / |h_k|$
    \EndFor
\EndFor
\State $I \gets softmax(I,\tau)$
\State \Return $I$
\end{algorithmic}
\label{algo}
\end{algorithm}

\begin{table*}[!htbp]
\vskip 0.3in

\centering
\fontsize{5.5}{6.5}\selectfont
\resizebox{\textwidth}{!}{%
\begin{tabular}{c|p{5cm}|c}
\toprule
&\multicolumn{1}{c|}{ \textbf{Question}} &\multicolumn{1}{c}{ \textbf{Answer}}  \\
\midrule[\heavyrulewidth]

\multirow{6}{*}{\rotatebox{90}{\textbf{TriviaQA}}}
&\multicolumn{1}{m{4cm}|}{Which American-born Sinclair won  the Nobel Prize for Literature in 1930?}&
Sinclair Lewis\\
\cmidrule{2-3}
&\multicolumn{1}{m{4cm}|}{Which musical featured the song 
Thank Heaven for Little Girls?}&
Gigi\\
\cmidrule{2-3}
&\multicolumn{1}{m{4cm}|}{What was the first movie 
western called?}&
Kit Carson\\

\midrule[\heavyrulewidth]
\multirow{3}{*}{\rotatebox{90}{\textbf{NaturalQA}}}
&\multicolumn{1}{m{4cm}|}{When did the eagles 
win last super bowl?}&2017\\
\cmidrule{2-3}
&\multicolumn{1}{m{4cm}|}{Who was the ruler of england in 1616?}&
James I\\
\cmidrule{2-3}
&\multicolumn{1}{m{4cm}|}{What is the hot coffee
mod in san andreas?}&a normally inaccessible mini-game\\

\midrule[\heavyrulewidth]
\multirow{5}{*}{\rotatebox{90}{\textbf{WebQA}}}
&\multicolumn{1}{m{4cm}|}{what character did
natalie portman play in star wars?}&Padmé Amidala\\
\cmidrule{2-3}
&\multicolumn{1}{m{4cm}|}{what country is the 
grand bahama island in?}&
Bahamas\\
\cmidrule{2-3}
&\multicolumn{1}{m{4cm}|}{where did saki live?}&
United Kingdom\\

\bottomrule
\end{tabular}}
\caption{Data samples from the datasets we use to evaluate UE methods: TriviaQA, NaturalQA, and WebQA. }
\label{tab:data_samples}
\end{table*}

\section{Experimental Details} \label{exp_details}

\noindent\textbf{Datasets.} We employ the validation split of the Natural Questions dataset, comprising 3610 samples. Following \citet{kuhn2023semantic}, a subset of 8000 QA pairs is selected from the validation split of the TriviaQA dataset. For WebQA, we combine its training and test splits to form a combined dataset of 6642 samples.

\bigskip

\noindent \textbf{Example Samples from Datasets.}
We provide data samples from the datasets we used in the evaluation of UE methods in Table~\ref{tab:data_samples}.

\bigskip

\noindent\textbf{Number of Sampling and Temperature.} Following previous work \cite{kuhn2023semantic}, we sampled 5 samples and used 0.5 as the temperature value for the results presented in Table \ref{tab:main_results}. 

\bigskip

\noindent \textbf{Generation Configurations.}
We use the Huggingface library's generate function for model generations. We set token ``.'' as eos\_token\_id which prevents model to generate long paragraphs to closed-book questions. We set num\_beams = 1 which corresponds to greedy decoding. 

\bigskip

\noindent \textbf{Computational Cost.}
We use 40 GB Nvidia A-100 GPUs for all the experiments. The total GPU-hours for Table \ref{tab:main_results} is approximately 400. Labeling of the data used for training of BERT-like importance model takes approximately 200 GPU-hours. Fine-tuning of BERT-like model on the importance dataset takes 7 GPU-hours. Due to expensive computational demands, all presented results are the output of a single run. 
\bigskip

\noindent \textbf{Prompts.}
We use the same 2-shot prompt for all of the models and the datasets for answer generation:

\begin{verbatim}
Answer these questions:
Question: What is the capital city of 
Australia?
Answer: The capital city of Australia is 
Canberra.
Question: Who painted the famous 
artwork "Starry Night"?
Answer: "Starry Night" was painted by 
Vincent van Gogh.
Question: {sample['question']}?
Answer: 
\end{verbatim}

To evaluate the correctness of the generated answer, we use gpt-3.5-turbo as the evaluator. The prompt for gpt-3.5-turbo is the following:
\begin{verbatim}

You will behave as a question-answer 
evaluator. I will give you a question, 
the ground truth of the question 
and a generated answer by a language 
model. You will output "correct" 
if the generated answer is correct 
regarding question and ground truth. 
Otherwise, output "false".
Question: {question}?, 
Ground Truth: {answer},
Generated Answer: {generation}
\end{verbatim}

\end{document}